%% file: main.tex
\documentclass[twoside]{article}

\usepackage[round]{natbib}
\usepackage{booktabs}       
\usepackage{graphicx}
\usepackage{amsmath}
\usepackage{amsfonts}
\usepackage{commath}
\usepackage{algpseudocode}
\usepackage{algorithm}
\usepackage{amssymb}
\usepackage{amsthm}
\usepackage{caption}
\usepackage{subcaption}
\usepackage[breaklinks = true]{hyperref}
\hypersetup{
    colorlinks,
    citecolor=blue,
    filecolor=pink,
    linkcolor=blue,
    urlcolor=blue
}

\algnewcommand{\require}[1]{%
  \State \textbf{Require: }\parbox[t]{.8\linewidth}{\raggedright #1}
}
\algnewcommand{\initialize}[1]{%
  \State \textbf{Initialize: }\parbox[t]{.8\linewidth}{\raggedright #1}
}

\theoremstyle{remark}

\theoremstyle{definition}
\newtheorem{definition}{Definition}

\usepackage[accepted]{aistats2023}

\begin{document}



\twocolumn[

\aistatstitle{Adversarial Learning Networks: Source-free Unsupervised Domain Incremental Learning}

\aistatsauthor{ Abhinit Kumar Ambastha \And Leong Tze Yun}

\aistatsaddress{ National University of Singapore \And  National University of Singapore } ]

\input{abstract}
\input{introduction}
\input{related_works}
\input{our_approach}
\input{experiments_and_results}
\input{conclusion}

\bibliography{ref}

\end{document}

%% file: abstract.tex
\begin{abstract}
This work presents an approach for incrementally updating deep neural network (DNN) models in a non-stationary environment. DNN models are sensitive to changes in input data distribution, which limits their application to problem settings with stationary input datasets. In a non-stationary environment, updating a DNN model requires parameter re-training or model fine-tuning. We propose an unsupervised source-free method to update DNN classification models. The contributions of this work are two-fold. First, we use trainable Gaussian prototypes to generate representative samples for future iterations; second, using unsupervised domain adaptation, we incrementally adapt the existing model using unlabelled data. Unlike existing methods, our approach can update a DNN model incrementally for non-stationary source and target tasks without storing past training data. We evaluated our work on incremental sentiment prediction and incremental disease prediction applications and compared our approach to state-of-the-art continual learning, domain adaptation, and ensemble learning methods. Our results show that our approach achieved improved performance compared to existing incremental learning methods. We observe minimal forgetting of past knowledge over many iterations, which can help us develop unsupervised self-learning systems.

\end{abstract}

%% file: introduction.tex
\section{Introduction}
Deep neural networks (DNN) have shown exceptional performance in several practical classification problems by approximating mapping functions between input data and output classes. However, they fail to generalize well in non-stationary environments where the labeling function changes continually due to shifts in the feature distribution of the input data \citep{mcgaughey2016understanding,pmlr-v119-chan20a,ioffe2015batch,bickel2009discriminative}. This is referred to as domain incremental learning problem setting and limits the applicability of DNNs to stationary domains with sizeable labeled training datasets. In order to minimize model degradation in this problem setting, we need to retrain the model every time we have an updated dataset. This process needs us to store previously observed training data and continuously label new incoming data \citep{hoffman2012discovering,srivastava2021continual,wulfmeier2018incremental}.
Continual learning methods and domain adaptation methods have been proposed in recent literature to address the problem of updating DNN models by adding additional network parameters to learn new mapping functions or store representative samples from past training data  \citep{li2017learning,hoffman2014continuous,zenke2017continual,srivastava2021continual}. Even though they minimize the forgetting of previously learned knowledge, they require future increments to be labeled and fail to address the issue of continuously annotating incoming data. Unsupervised domain adaptation methods overcome this limitation of labeling new data \citep{ganin2016domain,zhao2019multi,zhao2018adversarial,churamani2021domain}. However, they still require storing past labeled training data.

This work presents a source-free unsupervised domain incremental learning approach -- \textbf{adversarial learning network (ALeN)}. The main contribution of this work is that it enables incrementally updating a DNN model using unlabelled data without storing past training data. Our work is motivated by adversarial domain adaptation \citep{ganin2016domain} and class incremental domain adaptation \citep{kundu2020class}. \cite{ganin2016domain} presented an approach to carry out unsupervised adversarial domain adaptation at a fixed time-point which assumes stationary source and target tasks. Our work is an incremental extension of the approach and allows adapting a base model for a non-stationary target task. Our problem formulation helps position the incremental learning problem setting as a continual domain adaptation problem by assuming the incremental input as a single non-stationary dataset. \cite{kundu2020class} present a source-free approach for class incremental learning (learning additional classes from target data without storing source domain data). Their approach uses domain adaptation to retain source model knowledge but does not address incremental learning. We extend their approach to non-stationary input distributions. We use prototype networks to generate past training data representatives; this alleviates the need for storing past training samples and reduces the memory complexity of our approach.

\begin{figure*}[h]
    \centering
    \includegraphics[width=\textwidth]{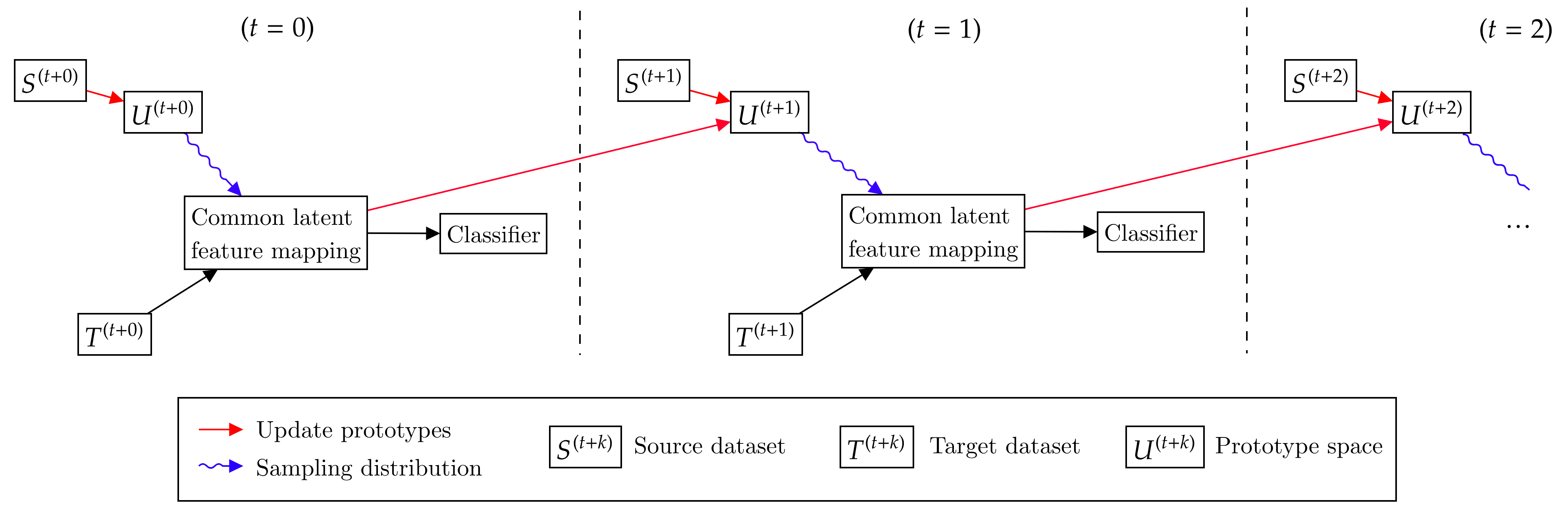}
    \caption{Generic architecture for the proposed unsupervised domain incremental learning method}
    \label{fig:overview}
\end{figure*}

We depict our method in Figure \ref{fig:overview}. Our method is divided in two stages -- \textit{base classifier learning} and \textit{incremental learning stage}. We train a base classifier for a supervised source task in the first stage. We learn the cluster prototypes from the feature space of this base classifier. We update the cluster prototypes using fine-tuning for a new source dataset every time. In the second stage, we use unsupervised domain adaptation to adapt the base classifier for a target task. We update the target task model using domain adaptation every time for incremental target dataset updates.

%% file: related_works.tex
\section{RELATED WORK}

Existing methods in the literature achieve incremental updates for DNN models by either learning a domain invariant feature mapping space or extending the model parameter space. In this section, we briefly study existing works for domain incremental learning.

\textbf{Unsupervised domain adaptation:} \cite{ganin2016domain} propose an adversarial domain adaptation approach to minimize domain discrepancy. They provide an adversarial unsupervised domain adaptation approach (DANN) to learn a common model for labeled source domain and unlabeled target domain datasets. Adversarial domain adaptation methods learn a domain in-variant feature space. \cite{zhao2018adversarial} further extend this approach as multi-source domain adaptation (MSDA) to learn a more generalized base source model and ensure that the domain discrepancy with the incoming target data would be lower. Discrepancy metrics are sensitive to feature representation \citep{hoffman2012discovering,blitzer2008learning,tzeng2017adversarial} and require hand-crafting source and target features. Although these methods update a model using unlabelled data, they are not incremental and require representative memory. \cite{kim2021domain} and \cite{yang2021generalized} present domain adaptation methods and do not address the problem setting with source and target domain data available incrementally. \cite{kim2021domain} use class-confident target domain samples to pseudo-label target domain data using a pre-trained source model. This can lead to negative training. We address negative training in our work using out-of-distribution samples and derive class prototypes from source-domain feature space. The proposed method cannot accommodate baseline source model updates. The approach requires re-training from scratch if the source domain distribution changes. Our approach can incrementally update the model using labeled source and unlabeled target domain data.

\textbf{Continual learning:} In contrast to domain adaptation methods, continual learning methods consider data from different domains sampled from a single non-stationary distribution. \cite{hoffman2014continuous} provide a continual adaptation (CMA) approach which learns a low dimensional embedding subspace for incoming non-stationary target data. The authors update a parametric kernel as the target domain distribution evolves\footnote{Kernel-based methods define a multi-kernel feature transformation function but are limited to pre-defined kernels due to symmetry and positive-definite constraints}. Also, the method assumes the sources are closely related to the target task. We relax this assumption in our method. Finally, they use a geodesic kernel to map target instances to the source domain. A limitation of this approach is that it does not address the possibility that an evolving source domain distribution requires re-training the source model using stored source domain data. Also, using a kernel-based approach is computationally intensive if the target dataset size is large, which limits the scalability of the approach.

\cite{castro2018end} propose a DNN-based continual learning approach to address the abovementioned limitations. The authors trained a joint feature extractor using stored labeled source and target domain data. This partially resolves the resource constraints in an incremental learning problem setting, but the requirement of labeled data makes it unsuitable for unsupervised domain incremental learning. \cite{li2017learning} provide a continual learning approach that expands the network parametric space to learn new classes and feature spaces using distillation loss. The method achieves low catastrophic forgetting but at the cost of using labeled source and target data.

\textbf{Ensemble learning:} ensemble learning approaches have been proposed in the literature to learn an incremental model using boosting and bagging approaches \citep{muhlbaier2008learn, polikar2001learn, muhlbaier2004learn, polikar2010learn, ditzler2010incremental}. \cite{elwell2011incremental} propose a boosting-based incremental learning algorithm that uses model weighting to adapt for non-stationary target data. The algorithm allocates higher weight to classifiers capable of identifying previously unseen instances while penalizing the other classifiers in the ensemble. The approach requires labeled source and target domain data like previous works.

A significant limitation of boosting-based incremental learning algorithms is the time complexity of training. Weights for all previous data sample errors are re-calculated in every iteration, which makes the approach unsuitable for large datasets and frequent incremental updates.

%% file: our_approach.tex
\section{OUR APPROACH}
\begin{figure}[h]
    \begin{center}
    \includegraphics[width=0.47\textwidth]{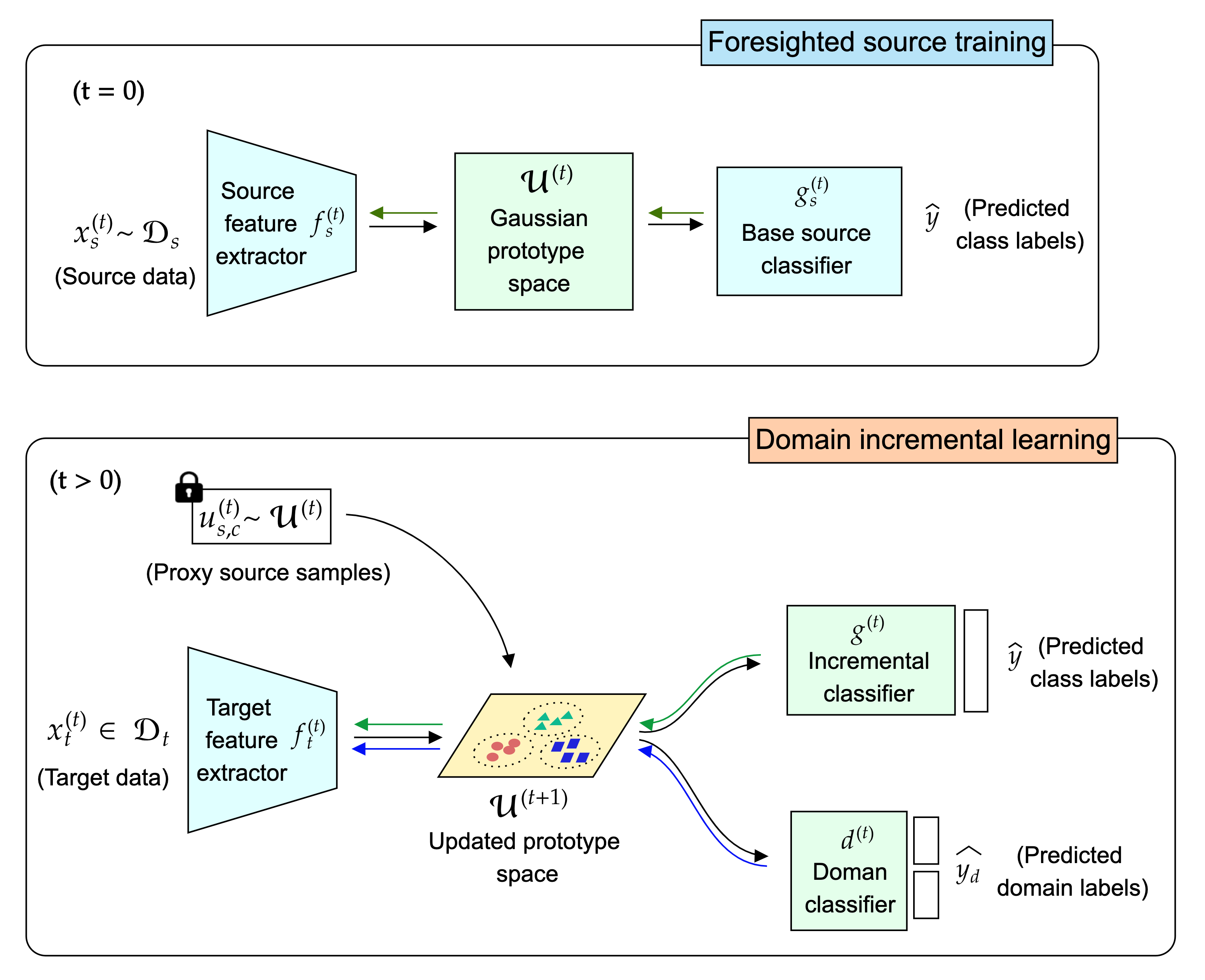}    
    \end{center}
    \caption{ALeN architecture: Architecture of the proposed domain incremental learning approach.}
    \label{fig:method}
\end{figure}
This section provides the details of ALeN (our proposed approach). Our method has two key components -- a knowledge representation mechanism for past knowledge replay and an incremental learning algorithm to learn an updated target task model.
Figure \ref{fig:method} shows the architecture of the proposed approach. The base model network includes two parts -- a feature extractor and a classifier. The feature extractor network learns a latent feature space from the input data which minimizes the classification loss. The classifier network maps the latent feature space to the label space. 

By learning a generative model for the latent features extracted from source domain data, we can generate representatives of the training data for the base model. We learn a Gaussian estimate for each class-wise source domain feature distribution.
Clustering methods use class-wise feature moments to define membership criteria for a given cluster. We estimate class-wise posterior distributions of the input data given a feature mapping network \citep{snell2017prototypical}. We use the Gaussian prototypes as \textit{guides} to generate class-wise data representatives.

\begin{definition}{\textbf{Gaussian prototypes:}}
The Gaussian prototypes for a given class $c$ are defined as a multi-variate Gaussian prior for each class $c$ in the latent space $\mathcal{U}$. It is given by $\mathcal{P}_s^c = \mathcal{N}(\mu_s^c,\Sigma_s^c)$, where $\mu_s^c$ and $\Sigma_s^c$ denote the mean and covariance of features $f_s(x_s)$, where $x_s \in [c]$.
\label{gaussian_prototypes}
\end{definition}

\subsection{Foresighted source training}

\begin{table*}[!ht]
\centering
\begin{tabular}{@{}lllllllll@{}}
\toprule
\multicolumn{3}{c}{\textbf{feature extractor ($f_s,f_t$)}} & \multicolumn{3}{c}{\textbf{Classifier ($g,g_s$)}} & \multicolumn{3}{c}{\textbf{Discriminator ($d$)}} \\ \midrule
ResNet-50 (till avg. pool layer) & 2048 &     & Input & 256  & -   & Input & 256 & -   \\
FC        & 1024 & ELU & FC    & 64   & ELU & FC    & 64  & ELU \\
BN        & -    &     & FC    & $|C_s+1|$ & -   & FC    & 2   & -   \\
FC        & 256  & ELU &       &      &     &       &     &     \\
FC        & 256  & ELU &       &      &     &       &     &     \\
BN        & -    &     &       &      &     &       &     &     \\ \bottomrule
\end{tabular}%
\caption{Network architecture for ALeN. $C_s$ is the number of classes.}
\label{tab:my-table}
\end{table*}

This section describes the foresighted learning stage. This stage aims to identify tight class-wise clusters in feature posterior distribution using Gaussian estimation. Algorithm \ref{foresighted_learning} describes the process of foresighted training and learning the source domain prototypes.

\begin{algorithm}[!h]
\caption{Foresighted baseline model training}
\begin{algorithmic}[1]
\require{Source samples $\mathcal{D}_s$, model parameters $\theta_{f_s}, \theta_{g_s}$, batch size of the source samples $N_{src}$ and \textit{negative} samples $N_{neg}$}
\item[]
\Repeat \label{1epoch_start}
    \State Obtain a mini-batch of source samples $S_s = \{\mathbf{x}_s,y_s) \sim \mathcal{D}_s\}$
    \State $\theta_{f_s} \gets \theta_{f_s} + \delta_{\theta_{f_s},S_s}$
\Until{reached the end of 1 epoch}\label{1epoch_end}
\item[]
\For {$c \in \mathcal{C}_s$} \label{proto_start_c}
    \State $\mathcal{D}_s^c \gets \{(\mathbf{x}_s,y_s):(\mathbf{x}_s,y_s) \in \mathcal{D}_s, y_s = c\}$
    \State $\mathbf{\mu}_s^c \gets \text{mean}_{(\mathbf{x}_s,y_s) \in \mathcal{D}_s^c}(f_s(\mathbf{x}_s))$
    \State $\mathbf{\Sigma}_s^c \gets \text{cov}_{(\mathbf{x}_s,y_s) \in \mathcal{D}_s^c}(f_s(\mathbf{x}_s))$
    \State $\mathcal{P}_s^c \gets \mathcal{N}(\mathbf{\mu}_s^c,\mathbf{\Sigma}_s^c)$
\EndFor\label{proto_end_c}
\State $\mathbf{\mu}_s \gets \text{mean}_{(\mathbf{x}_s,y_s) \in \mathcal{D}_s}(f_s(\mathbf{x}_s))$ \label{proto_start}
\State $\mathbf{\Sigma}_s \gets \text{cov}_{(\mathbf{x}_s,y_s) \in \mathcal{D}_s}(f_s(\mathbf{x}_s))$
\State $\mathcal{P}_s \gets \mathcal{N}(\mathbf{\mu}_s,\mathbf{\Sigma}_s)$ \label{proto_end}
\item[]
\State \textbf{Loss} $\gets [\mathcal{L}_{s1},\mathcal{L}_{s2}]$\label{loss_start}
\State \textbf{Opt} $\gets [\text{\textit{Adam}}_{\{f_s\}},\text{\textit{Adam}}_{\{f_s,g_s\}}]$
\State $\Phi \gets [\{\theta_{f_s}\},\{\theta_{f_s,g_s}\}]$
\State \textit{iter} $\gets 0$\label{loss_end}
\item[]
\Repeat
    \State iter $\gets$ iter $+ 1$
    \State cur $\gets$ iter mod $2$
    \item[]
    \State $S_s \gets \{(\mathbf{x}_s,y_s) \sim \mathcal{D}_s\}$ \label{neg_start}
    \State $S_n \gets \text{IdentifyNegativeSamples}(\mathcal{P}_s,\{\mathcal{P}_s^c:c\in \mathcal{C}_s\},N_{neg})$
    \For {$(\mathbf{x}_s,y_s) \in S_s$}
        \State $\mathbf{y}_s \gets g_s \circ f_s(\mathbf{x}_s)$
    \EndFor
    \For {$(\mathbf{u}_n,y_n) \in S_n$}
        \State $\mathbf{y}_n \gets g_s(\mathbf{u}_n)$
    \EndFor \label{neg_end}
    \State $\Phi[\text{cur}] \gets \Phi[\text{cur}] + \textbf{Opt}[\text{cur}](-\delta_{\Phi[\text{cur}],\textbf{Loss}[\text{cur}]})$
    \If{\textit{reached the end of an epoch}}
        \State Recalculate Gaussian prototypes following lines 6-14
    \EndIf
\Until{Convergence}
\end{algorithmic}
\label{foresighted_learning}
\end{algorithm}
We denote the feature extractor function as $f_s$ and the classifier function as $g_s$, which maps the feature extractor output to a $|C_{s}+1|$-class label space (where $C_s$ is the source task label set size). The latent space is denoted by $\mathcal{U}$. We minimize cross-entropy loss ($l_{ce}$) to learn $g_s$.
\begin{equation}
    l_{ce} = \mathop{\mathbb{E}}_{(x_s,y_s)\sim \mathcal{D}^{(t)}_{s}} l_{ce}(g_s \cdot f_s (x_s), y_s)
    \label{vanilla_ce}
\end{equation}
Minimizing $l_{ce}$ alone has been shown to bias the base model to source domain characteristics \citep{kundu2020class}. Cross-entropy loss ensures discriminative decision boundaries in the latent feature space but leads to over-confident predictions. In order to generate representative samples for source distribution for future iterations, we minimize category bias by penalizing over-confident prediction. We achieve this by identifying out-of-distribution samples (OOD samples). We reduce the number of negative samples (low-confidence source domain samples) by training the classifier to identify a $|C_{s}+1|$-class for OOD samples. The negative samples are identified based on the learned class-wise prototypes $u_s^c \sim \mathcal{P}_s^c, c\in C_s$ using a $k-\sigma$ confidence interval (line~\ref{neg_start}-\ref{neg_end}) (we identified $k=3$ using sensitivity analysis). 

We use a class separability objective $\mathcal{L}_{s1}$ to enforce the class-wise features to attain higher affinity to the class-wise prototypes
\begin{equation}
    \mathcal{L}_{s1}: \mathop{\mathbb{E}}_{(x_s,y_s)\sim \mathcal{D}^{(t)}_{s}} - \log \left[ \frac{\exp(\mathcal{P}_s^{y_s}(f_s^{(t)}(u_s)))}{\sum_{c\in C_s}\exp(\mathcal{P}_s^{c}(f_s^{(t)}(u_s)))} \right]
\end{equation}
To update the prototypes at the end of the training, we use a class separability approach to learn a stationary source distribution space ($\mathcal{U}-space$) and store the guides for the upcoming incremental phase. These guides are used to sample pseudo-source domain instances in the incremental learning stage and apply pseudo labels to them.
\begin{equation}
    \mathcal{L}_s = \mathcal{L}_{s1} + \mathcal{L}_{s2}
\end{equation}
\begin{equation}
\begin{split}
    \mathcal{L}_{s2}: \mathop{\mathbb{E}}_{(x_s,y_s)\sim \mathcal{D}^{(t)}_{s}} l_{ce}(g_s \cdot f_s(x_s), y_s) \\
    + \mathop{\mathbb{E}}_{(u_n,y_n)\sim \bar{\mathcal{D}}^{(t)}_{s}} l_{ce}(g_s(u_n), y_n)
\end{split}
\end{equation}
\subsection{Domain incremental update algorithm}

In this section, we describe the domain incremental update stage of the proposed method. We use the learned prototypes and the unlabelled target domain data to incrementally updated base classifier for the target task. Algorithm~\ref{domain_incremental_algo} presents the proposed algorithm. 

\begin{algorithm}[!ht]
\caption{Domain Incremental Learning}
\begin{algorithmic}[1]
\require{Target samples $\mathcal{D}_t$, Gaussian Prototypes $\mathcal{P}_s^c$, model parameters $\theta_{f_s^{(t)}}, \theta_{g_s^{(t)}}, \theta_{f_t^{(t)}}, \theta_{g_t^{(t)}}, \theta_{d^{(t)}}$, Training sample size $N$}
\initialize{$\theta_{f_t^{(t)}} \gets \theta_{f_s^{(t)}}$, $\theta_{g_t^{(t)}} \gets \theta_{g_s^{(t)}}$} \label{dom:init}
\State \textbf{Loss} $\gets [\mathcal{L}_c,\mathcal{L}_{d}]$
\State \textbf{Opt} $\gets [\text{Adam}_{\{f_t^{(t)}\}},\text{Adam}_{\{d^{(t)}, f_t^{(t)}\}}]$
\State \textit{iter} $\gets 0$
\Repeat
    \State \textit{iter} $\gets$ \textit{iter}$+1$
    \item[]
    \For{$u_s^c \sim \mathcal{P}_s^c$}  \label{dom:source_training_start}
        \State $\hat{y} \gets g_t(u_s^c)$
        \State $\mathcal{L}_c \gets \mathcal{L}_c + l_{ce}(y_s,c)$
    \EndFor \label{dom:source_training_end}
    \For{$u_s^c \sim \mathcal{P}_s^c$, $\mathbf{x}_t \in \mathcal{D}_t$} \label{dom:adversarial_training_start}
        \State $v \gets f_t(\mathbf{x}_t)$
        \State $\hat{y_d} \gets d([u_s^c,v])$
        \State $\mathcal{L}_d \gets \mathcal{L}_d + l_{ce}(\hat{y_d},[0,1])$
    \EndFor \label{dom:adversarial_training_end}
    \State $\theta_{f_t^{(t)}} \gets \theta_{f_t^{(t)}} + \text{Adam}_{\{f_t^{(t)}\}}(-\nabla\frac{1}{N}\sum \mathcal{L}_c)$ \label{dom:classification_loss_update}
    \State $\theta_{d^{(t)}} \gets \theta_{d^{(t)}} + \text{Adam}_{\{d^{(t)}, f_t^{(t)}\}}(-\nabla\frac{1}{N}\sum \mathcal{L}_d)$ \label{dom:discriminator_loss_update}
    \State $\theta_{f_t^{(t)}} \gets \theta_{f_t^{(t)}} - \text{Adam}_{\{d^{(t)}, f_t^{(t)}\}}(-\nabla\frac{1}{N}\sum \mathcal{L}_d)$ \label{dom:adversarial_loss_update}
\Until{Convergence}
\end{algorithmic}
\label{domain_incremental_algo}
\end{algorithm}

The incremental learning network comprises three sub-modules -- feature extractor, classifier, and domain discriminator. The feature extractor and classifier use the pre-trained base model parameters (line \ref{dom:init}) and have the same architecture as the base model. In order to retain source domain knowledge during this stage, we sample the class-wise prototypes to obtain labelled source domain representatives (line~\ref{dom:source_training_end} - \ref{dom:source_training_end}). We carry out adversarial domain adaptation to learn a target model using unlabeled target domain data \citep{ganin2016domain, tzeng2017adversarial} and the sampled source domain representatives. The feature extractor learns a common latent feature mapping for both the source and target domain. Line~\ref{dom:adversarial_training_start}-\ref{dom:adversarial_training_end} shows the target domain sampling and feature extraction. 

We use a domain discriminator network $d^{(t)}$ to estimate the $H-$distance between the source and target domains. The target domain features and source domain samples are passed to the domain discriminator. The true labels for the domain discriminator are provided incorrectly to increase the domain confusion loss. A gradient reversal layer between the domain discriminator and the target feature extractor is used to adversarially train the feature extractor (line\ref{dom:discriminator_loss_update} and line~\ref{dom:adversarial_loss_update}). We use cross-entropy loss to update the base classifier (using only source domain representatives) to ensure the retention of previously learned decision boundaries (line~\ref{dom:classification_loss_update}). We update the feature extractor and domain discriminator until convergence. 

%% file: experiments_and_results.tex
\section{EXPERIMENTS AND RESULTS}

\textbf{Setup:} The experiments were written in Python, and PyTorch was used to implement the neural network architecture, gradient propagation, and the training loop. SpaCy was used to implement the NLP functions and preprocessing steps. We used a workstation with a Tesla V100 graphics card with 32GB GPU memory and 256GB RAM. 

\subsection{Incremental Amazon products review prediction}

We illustrate the performance of our proposed method using a sentiment classification task for Amazon product reviews \citep{he2016ups,mcauley2015image}. The task is to predict a given product rating based on an incoming user review of the product. The rating is an integer in the range $[1,5]$. We use the Amazon reviews dataset to construct an incremental learning problem for predicting product ratings for a category (target task) for which we only have access to unlabelled data. The remaining product review datasets are used as sources available to the training algorithm sequentially in a randomized order. In total, the dataset contains labeled review data from 25 categories (we ignore three categories with insufficient data). We extract 5000 samples from each category.

The relative positioning of word vectors encodes the semantic meaning of a sentence. Hence, even though two corpora may contain words from the same vocabulary, their semantic meanings can differ. A sentiment prediction model aims to infer the sentiment of a phrase by analyzing the joint distribution of the set of tokens in the sentence. Due to the extensive vocabulary size, estimating this joint distribution for a given corpus can be infeasible. Hence, it is desirable to be able to adapt an existing model trained on a large labeled corpus to a smaller or unlabelled corpus drawn from the same vocabulary. In this experiment, we aim to incrementally learn a model for an unlabelled corpus given multiple sequentially available labeled corpora (drawn from the same vocabulary).

We evaluated our method on the sentiment classification task on the Amazon reviews dataset \citep{mcauley2015image,he2016ups}, which consists of global vectors (GloVe) for product text reviews. We extracted tokens by removing duplicate and incomplete rows and removing stop words, punctuation, and alphanumeric words. We converted the resultant tokens to GloVe vectors \citep{pennington2014glove} of shape $(1,300)$ and computed a mean vector for the review. We processed a maximum of 5000 reviews for every product. We used the review rating as the label.

\begin{figure*}[!h]
    \centering
    \begin{subfigure}[t]{0.50\textwidth}
  	 	\begin{center}
		\includegraphics[width=\textwidth]{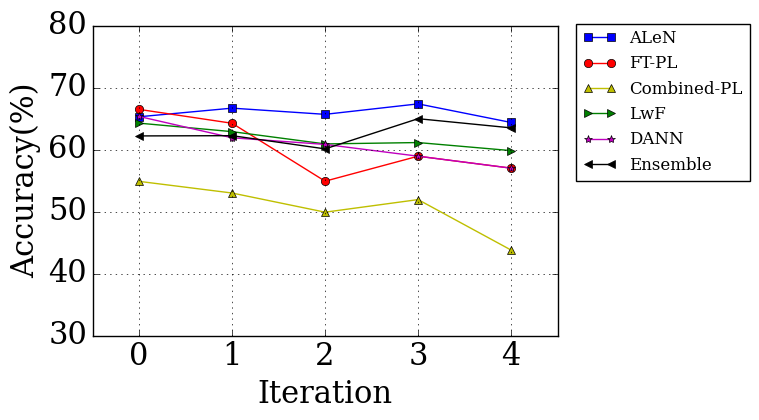}
		\caption{Electronics reviews target}
		\label{forgetting}
		\end{center}
    \end{subfigure}%
    \hfill
    \begin{subfigure}[t]{0.50\textwidth}
    	\begin{center}
		\includegraphics[width=\textwidth]{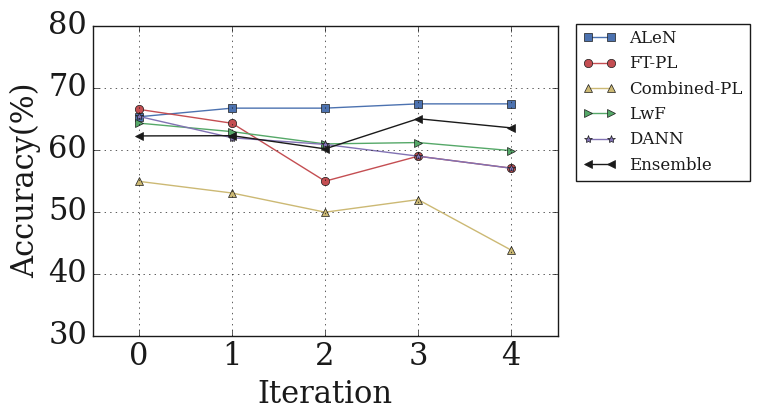} 
		\caption{Grocery reviews target}
		\label{forgetting}
		\end{center}
    \end{subfigure}%
    \vskip\baselineskip
    \begin{subfigure}[t]{0.50\textwidth}
    	\begin{center}
		\includegraphics[width=\textwidth]{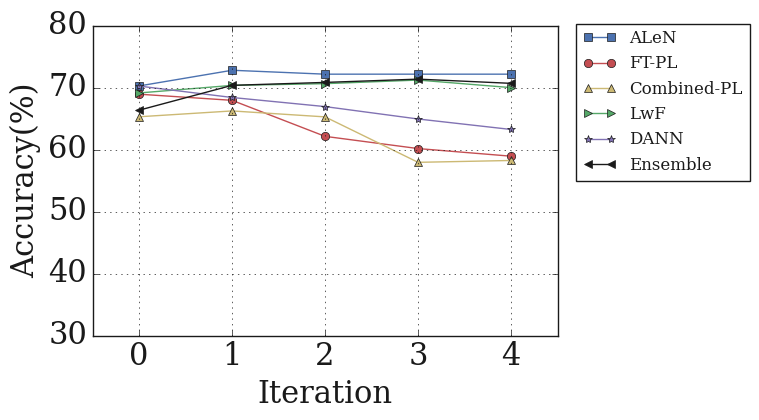} 
		\caption{Movies reviews target}
		\label{forgetting}
		\end{center}
    \end{subfigure}%
    \hfill
    \begin{subfigure}[t]{0.50\textwidth}
  	 	\begin{center}
		\includegraphics[width=\textwidth]{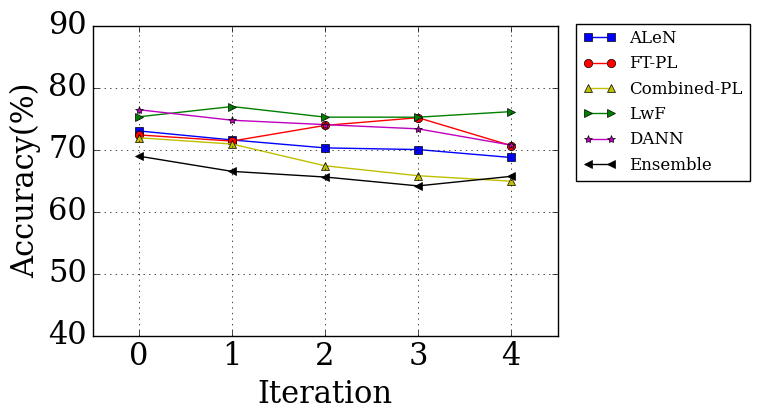} 
		\caption{Books reviews target}
		\label{forgetting}
		\end{center}
    \end{subfigure}%
    \caption{Comparison of related unsupervised incremental learning methods with the proposed method (ALeN). (a-d) show results for five increments for Electronics, Grocery, Movies, and Books review sentiment prediction target domains, respectively }
    \label{alen_accuracy_senti}
\end{figure*}

\begin{figure*}[!h]
    \centering
    \begin{subfigure}[t]{0.50\textwidth}
  	 	\begin{center}
		\includegraphics[width=\textwidth]{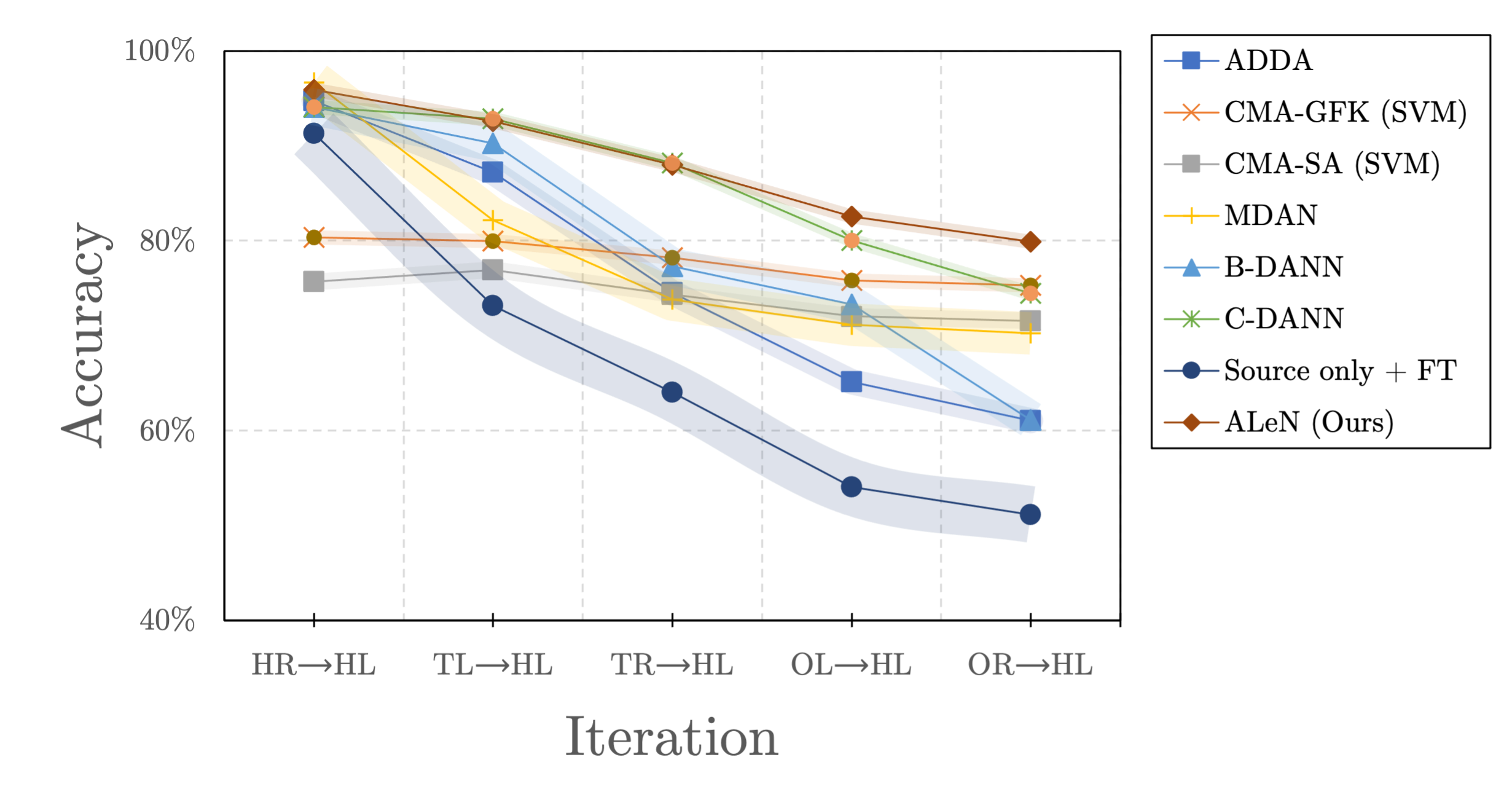}
		\caption{Hippocampal left (HL) target domain}
		\label{forgetting}
		\end{center}
    \end{subfigure}%
    \hfill
    \begin{subfigure}[t]{0.50\textwidth}
    	\begin{center}
		\includegraphics[width=\textwidth]{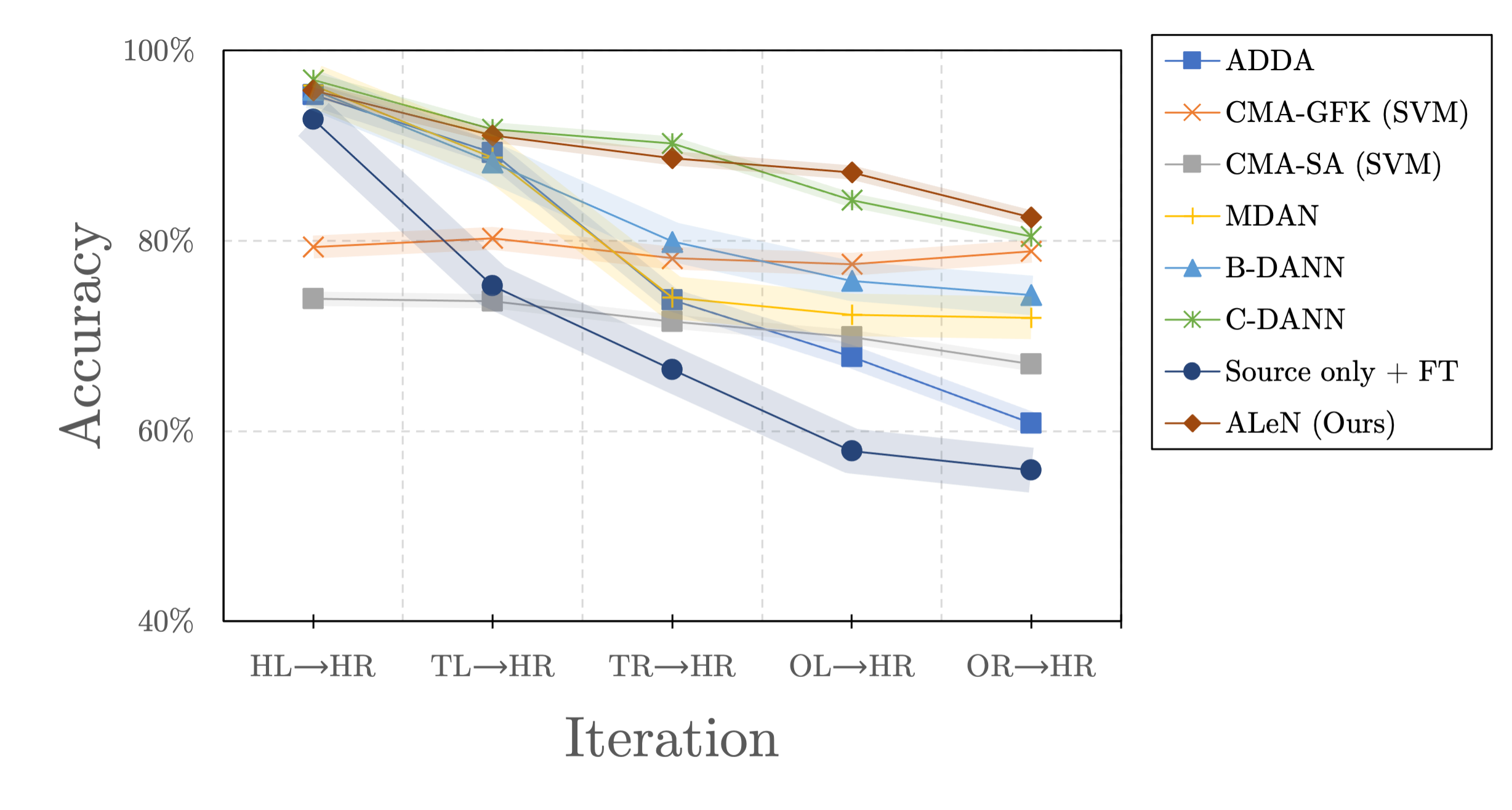}
		\caption{Hippocampal right (HR) target domain}
		\label{forgetting}
		\end{center}
    \end{subfigure}%
    \vskip\baselineskip
    \begin{subfigure}[t]{0.50\textwidth}
    	\begin{center}
		\includegraphics[width=\textwidth]{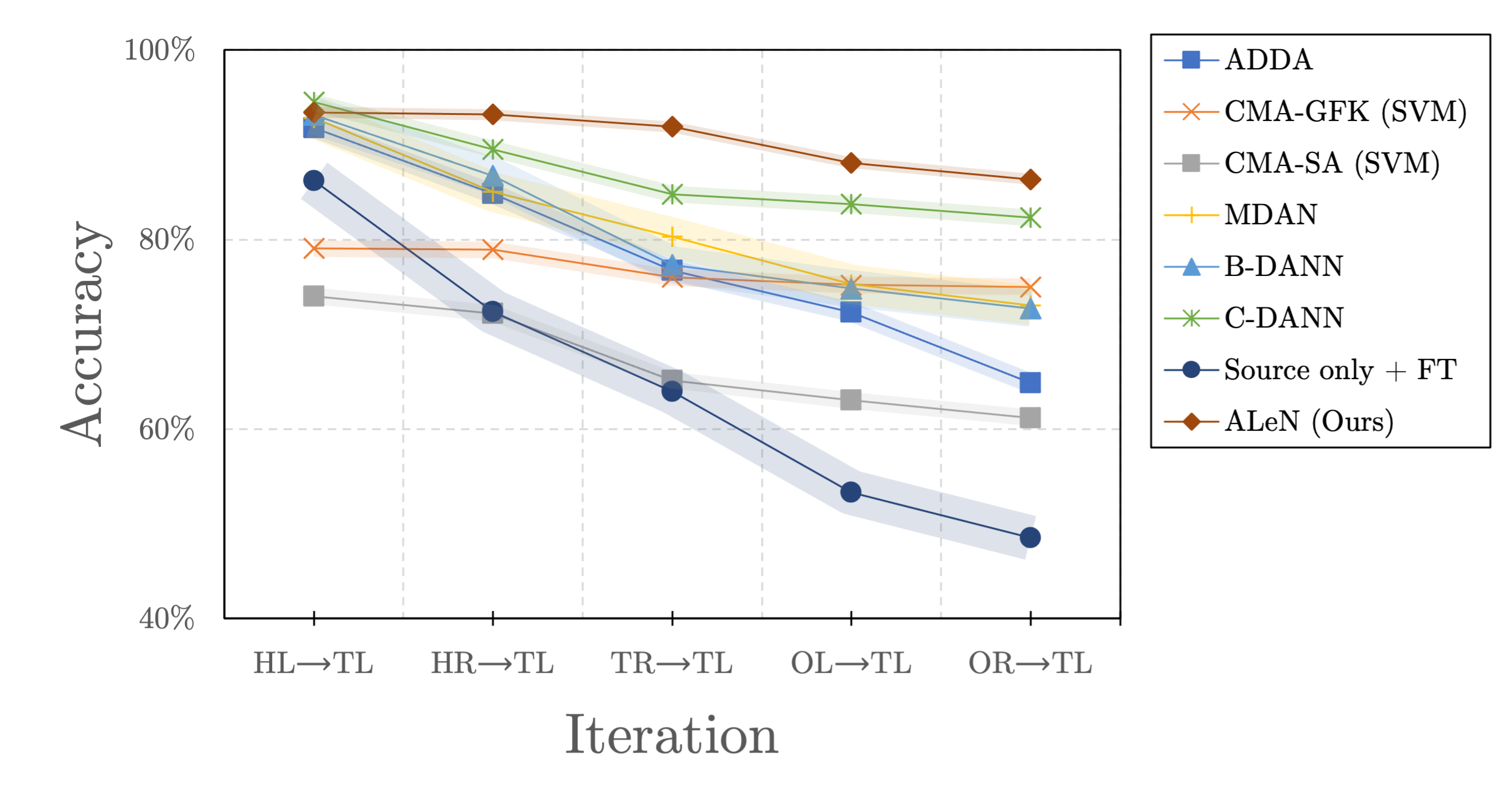}
		\caption{Temporal left (TL) target domain}
		\label{forgetting}
		\end{center}
    \end{subfigure}%
    \hfill
    \begin{subfigure}[t]{0.50\textwidth}
  	 	\begin{center}
		\includegraphics[width=\textwidth]{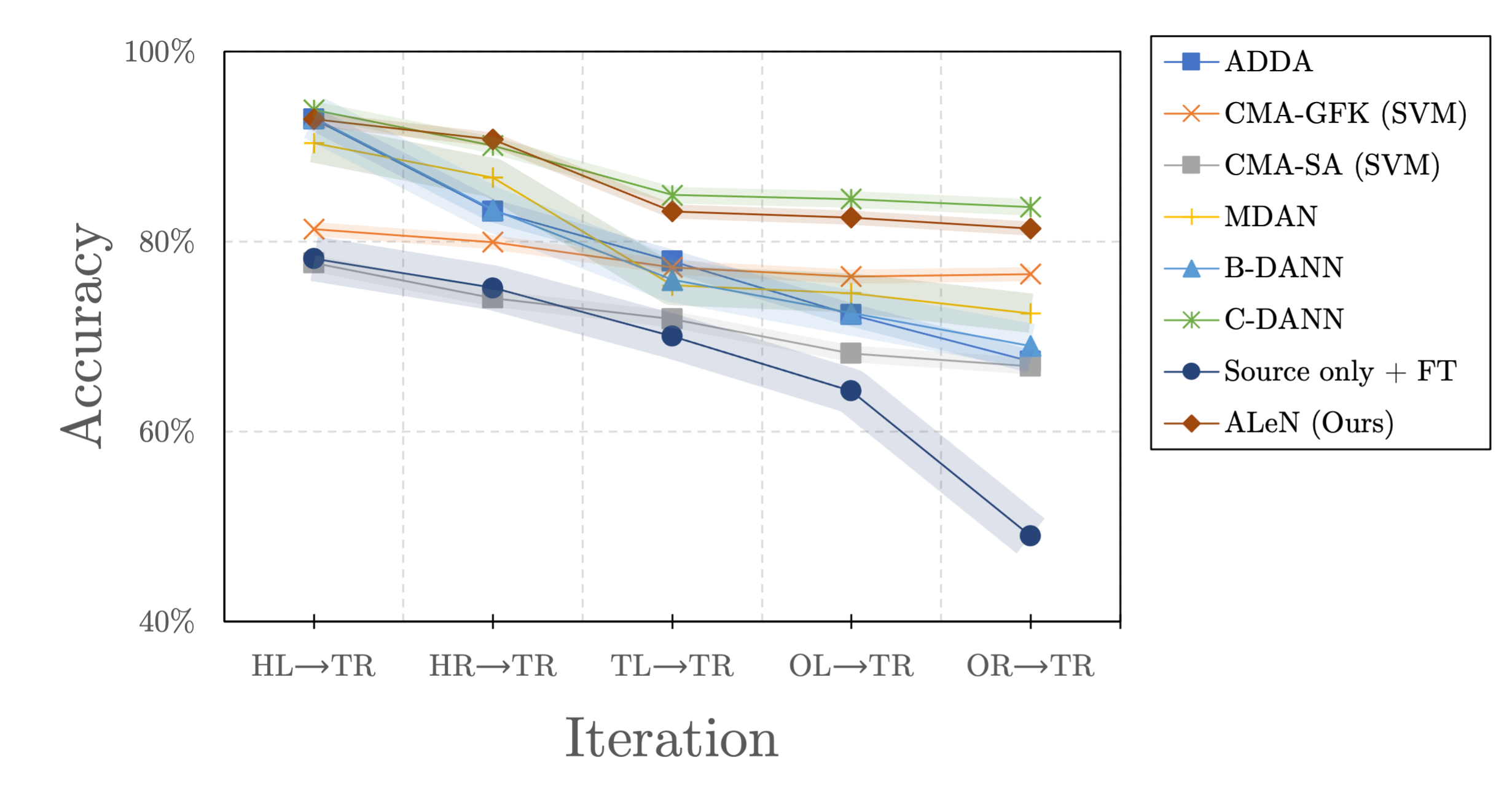}
		\caption{Temporal right (TR) target domain}
		\label{forgetting}
		\end{center}
    \end{subfigure}%
    \vskip\baselineskip
    \begin{subfigure}[t]{0.50\textwidth}
    	\begin{center}
		\includegraphics[width=\textwidth]{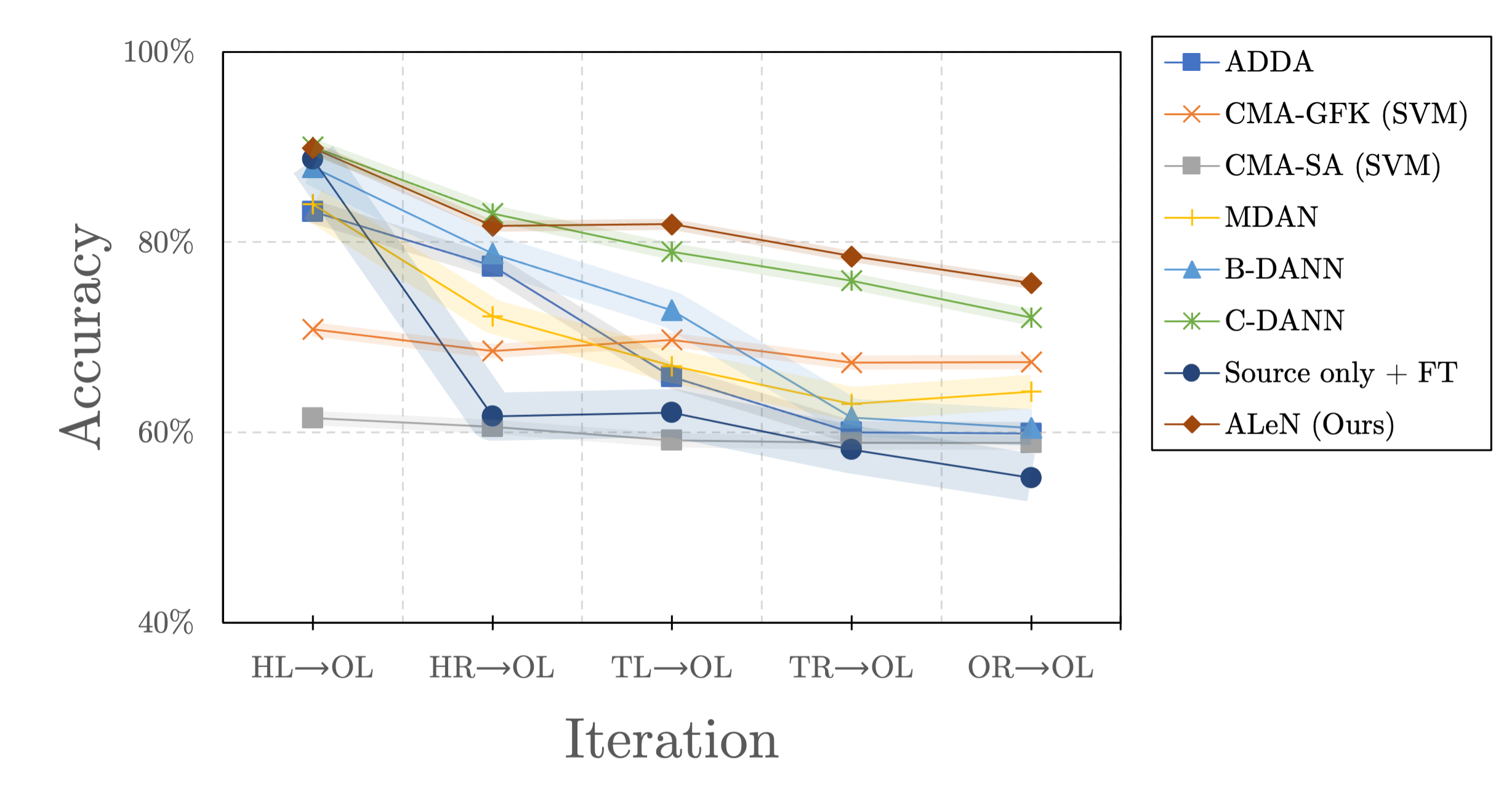}
		\caption{Occipital left (OL) target domain}
		\label{forgetting}
		\end{center}
    \end{subfigure}%
    \hfill
    \begin{subfigure}[t]{0.50\textwidth}
  	 	\begin{center}
		\includegraphics[width=\textwidth]{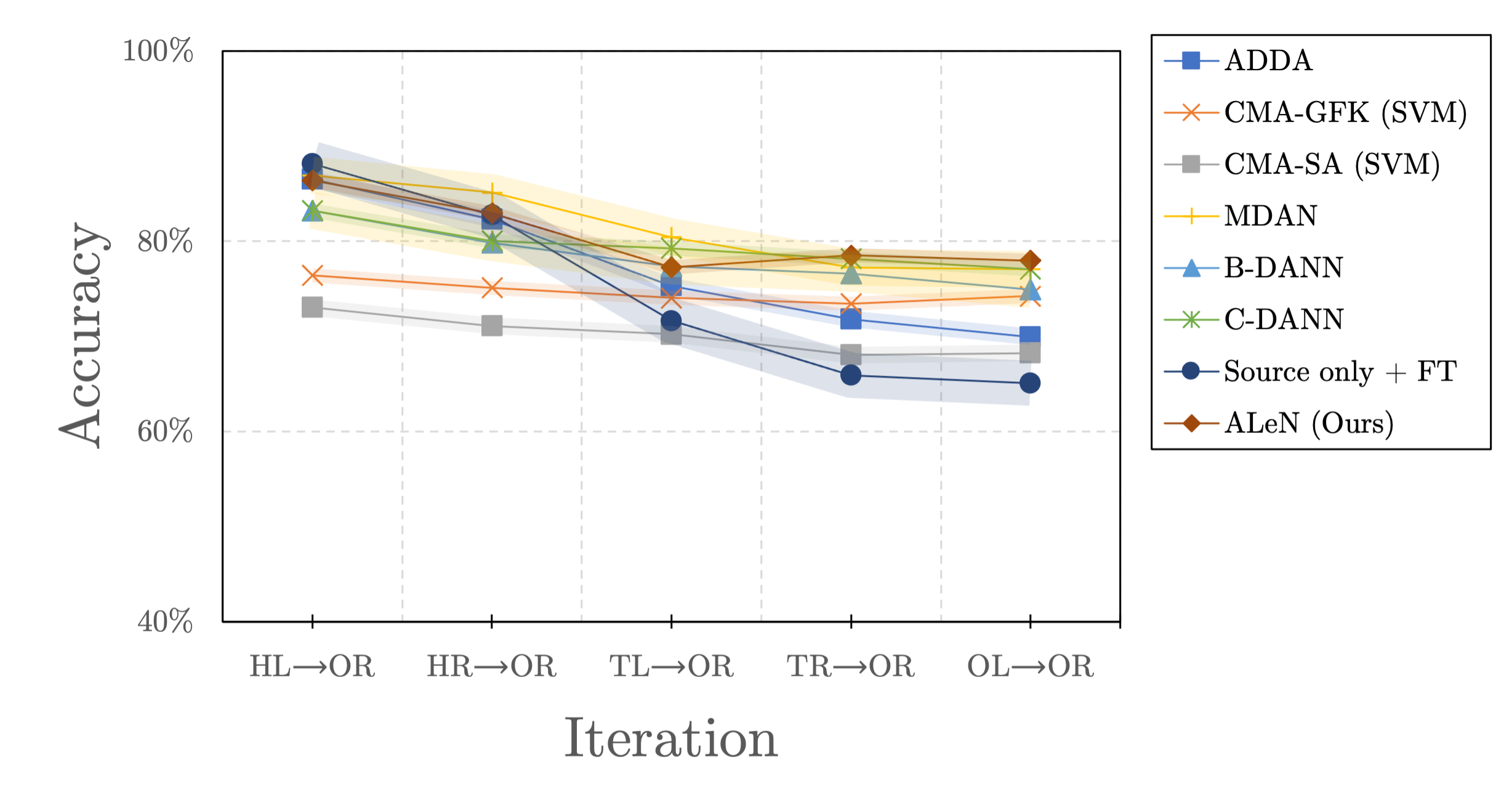} 
		\caption{Occipital right (OR) target domain}
		\label{forgetting}
		\end{center}
    \end{subfigure}%
    \caption{Comparison of related unsupervised incremental learning methods with the proposed method (ALeN). (a-f) show the overall accuracy for past domains for all iterations for different target domains.}
    \label{alen_accuracy_adni}
\end{figure*}

\begin{table*}[!h]
\centering
\resizebox{0.8\textwidth}{!}{%
{\renewcommand{\arraystretch}{1.5}
\begin{tabular}{{|l|l|l|l|l|l|l|l|l|}}
\hline
 &
  \multicolumn{2}{c|}{\textbf{ADDA}} &
  \multicolumn{2}{c|}{\textbf{CMA-GFK (SVM)}} &
  \multicolumn{2}{c|}{\textbf{CMA-SA (SVM)}} &
  \multicolumn{2}{c|}{\textbf{MDAN}} \\ \hline
 \textbf{Target} &
  \multicolumn{1}{c|}{ Avg. acc} &
  \multicolumn{1}{c|}{ Fgt(\%)} &
  \multicolumn{1}{c|}{ Avg. acc} &
  \multicolumn{1}{c|}{ Fgt(\%)} &
  \multicolumn{1}{c|}{ Avg. acc} &
  \multicolumn{1}{c|}{ Fgt(\%)} &
  \multicolumn{1}{c|}{ Avg. acc} &
  \multicolumn{1}{c|}{ Fgt(\%)} \\ \hline
Hippocampal L & 76.51\% & -0.83\% & 77.89\% & 3.48\%  & 74.09\% & 4.34\%  & 78.79\%          & 0.17\%            \\ \hline
Hippocampal R & 77.41\% & 0.00\%  & 78.86\% & 1.76\%  & 71.22\% & 0.10\%  & 80.65\%          & 0.00\%            \\ \hline
Temporal L    & 78.10\% & -4.66\% & 76.83\% & 2.89\%  & 67.11\% & 3.19\%  & 81.28\%          & -3.14\%           \\ \hline
Temporal R    & 78.74\% & 0.00\%  & 78.28\% & 1.31\%  & 71.73\% & 5.22\%  & 79.90\%          & 0.00\%            \\ \hline
Occipital L   & 69.29\% & 0.00\%  & 68.77\% & 1.21\%  & 59.81\% & 2.40\%  & 70.10\%          & 0.00\%            \\ \hline
Occipital R   & 77.14\% & -1.19\% & 74.62\% & -0.95\% & 70.10\% & 9.65\%  & \textbf{81.32\% }         & -2.98\%           \\ \hline
&
  \multicolumn{2}{c|}{ \textbf{B-DANN}} &
  \multicolumn{2}{c|}{ \textbf{C-DANN}} &
  \multicolumn{2}{c|}{ \textbf{FT}} &
  \multicolumn{2}{c|}{ \textbf{ALeN}} \\ \hline
 \textbf{Target} &
  \multicolumn{1}{c|}{ Avg. acc} &
  \multicolumn{1}{c|}{ Fgt(\%)} &
  \multicolumn{1}{c|}{ Avg. acc} &
  \multicolumn{1}{c|}{ Fgt(\%)} &
  \multicolumn{1}{c|}{ Avg. acc} &
  \multicolumn{1}{c|}{ Fgt(\%)} &
  \multicolumn{1}{c|}{ \textbf{Avg. acc}} &
  \multicolumn{1}{c|}{ \textbf{Fgt(\%)}} \\ \hline
Hippocampal L & 79.21\% & 1.42\%  & 85.92\% & -0.73\% & 66.71\% & 0.00\%  & \textbf{87.88\%} & \textbf{-2.90\%}  \\ \hline
Hippocampal R & 82.81\% & -0.91\% & 88.73\% & -7.48\% & 69.67\% & -0.42\% & \textbf{89.46\%} & \textbf{0.70\%}   \\ \hline
Temporal L    & 80.94\% & -1.72\% & 86.96\% & -0.13\% & 64.87\% & -0.42\% & \textbf{90.79\%} & \textbf{-0.17\%}  \\ \hline
Temporal R    & 78.76\% & 3.74\%  & 87.39\% & -8.36\% & 67.32\% & -0.78\% & \textbf{86.73\%} & \textbf{-1.25\%}  \\ \hline
Occipital L   & 72.32\% & 4.82\%  & \textbf{80.02\%} & -0.44\% & 65.19\% & -0.35\% & 74.74\% & \textbf{-3.39\%} \\ \hline
Occipital R   & 78.37\% & 2.01\%  & 79.52\% & -7.12\% & 74.67\% & 0.00\%  & 80.58\% & \textbf{-5.11\%}  \\ \hline
\end{tabular}}%
}
\caption{A comparison of our method (ALeN) with unsupervised adversarial domain adaptation method (ADDA), unsupervised continual learning methods (CMA-GFK, CMA-SA), unsupervised multi-task learning methods (MDAN, B-DANN, C-DANN) and fine-tuning (FT) using only labeled source data. \textit{Avg. acc (\%)} is the average overall accuracy and \textit{Fgt(\%)} is the forgetting at the end of all iterations. We average the results over 100 runs.}
\label{adni_table}
\end{table*}

We compare the final predictive accuracy after 24 iterations in an incremental setting for categories of the amazon reviews dataset. We set each category as the target dataset and the rest of the datasets. The first iteration ($t+0$) is the baseline model trained using the first source dataset picked randomly and the selected target dataset. We split the target dataset into training and test data using stratified sampling to ensure robust testing. We remove the labels for the training target data. 

The results (Figure \ref{alen_accuracy_senti}) clearly show that modeling the task as an incremental learning task leads to better target task performance for most categories. We observe that compared methods perform better or equally good for digital music, groceries, and electronics categories, which we observed was due to near normal distribution of the label set, with the majority of labels being 2 or 3. 

\subsection{Incremental learning for Alzheimer's disease prediction}

Alzheimer's disease has been the focus of many disease prediction works due to the availability of labeled data from the Alzheimer's Disease Neuroimaging Initiative (ADNI) database (adni.loni.usc.edu). The dataset was collected from the US and Canadian populations. This dataset has been the source for multiple studies and disease prediction models \citep{ dimitriadis2018random, parisot2018disease, mueller2005ways, hansson2018csf, eskildsen2013prediction}. Models trained on the ADNI dataset have poor performance when applied to real-world data from a population other than US and Canada. We overcome this bottleneck by formulating an incremental learning scenario, with available labeled datasets as sources and the unlabelled target population dataset as the target task.

In the following experiments, we use brain MRI imaging data obtained from ADNI and the Australian Imaging, Biomarker \& Lifestyle Flagship Study of Ageing (AIBL) datasets. ADNI and AIBL are multi-site longitudinal studies aimed at understanding the progression of Alzheimer's disease. The datasets consist of clinical, imaging, genetic, and biomarker data. We are interested in predictive models for the early detection of dementia; hence we select the screening stage MRI images for training.

Alzheimer's disease progression symptoms include atrophied brain regions, especially the Hippocampus and Temporal regions. We used voxel-based morphometry (VBM) \citep{ashburner2000voxel} to extract regional volumes for different regions of interest in the brain. We extract 112 regions of interest from the preprocessed gray matter images using the automated anatomical labeling atlas (AAL-MNI152 template) to create volumetric datasets for individual Regions of interest (ROIs).

We use ten dataset replicates for this experiment. We train the models with ten random parameter initializations. We use an $80/10/10$ split for training, testing, and validation. We present the results in Figure \ref{alen_accuracy_adni}, showing that our proposed approach (ALeN) can incrementally update the model with minimal forgetting for previously observed domains.

Table \ref{adni_table} shows a comparison of the average accuracy and forgetting of the existing methods and the proposed method. We observe that the unsupervised continual learning methods (CMA-GFK and CMA-SA) were able to learn an incremental hypothesis but failed to achieve a high prediction accuracy. The methods use a domain alignment approach which assumes a linear mapping between a low-dimension source and target embedding. Embeddings generated using feature reduction methods such as principal component analysis (PCA) might not be able to learn very efficient principal components if the features do not have much variance. Also, we observe that the support vector machine (SVM) classifier has a low accuracy on the task, which reduces the overall accuracy of the approach. 

The adversarial domain adaptation methods (MDAN and B-DANN) select the best feature mapping for the source domain. These methods' target domain accuracy is high, but they fail to learn from the source domain data incrementally. When we apply this feature mapping function to the target domain, we realize that if the model has a low error on the source domain, it stops updating the feature extractor for future source domains.

C-DANN is a variant of the adversarial domain adaptation network in which we provide a representational memory. This memory is used to store past data samples but has a limited size. This emulates an incremental learning problem setting where we store past samples. We observe that the method could minimize forgetting using the stored samples. However, due to the limited size of the memory, forgetting occurs as the number of increments increases.

\subsection{Ablation study and sensitivity analysis}

We present a sensitivity analysis and an ablation study for the proposed method. We illustrate the effects of the analysis on the automated disease diagnosis task for predicting Alzheimer's disease using a non-stationary source domain (simulated using regions Hippocampus Right, Temporal left and right, and occipital left and right) and non-stationary target domain data (sampled from Hippocampus Left).

\textbf{Effectiveness of Gaussian prototypes and OOD samples:} We analyzed the sensitivity of the hyper-parameter $k$ used for carrying out negative sample identification in the foresighted learning algorithm (algorithm \ref{foresighted_learning}). We use a 3-$\sigma$ confidence interval to pseudo-label the $|C_{s}+1|^{th}$-class OOD samples following the empirical rule for Gaussian distributions. In order to verify that our Gaussian estimates are accurate, we empirically tested the efficiency of the assumed confidence interval. We observed that 3-$\sigma$ provided the maximum predictive accuracy and best captured the source distribution characteristics. By setting $k$ to 5, we can set the minimum number of negative samples, which can be considered an ablation of OOD samples. We observe the average accuracy for the target task to be 77.20\% compared to 89.46\% when using a 3-$\sigma$ confidence interval.
    
\textbf{Effect of balancing source and target unlabelled data:} We used a balanced source ($N_{src}$) and target ($N_{neg}$) domain dataset to train our baseline model. We test the robustness of our model to imbalanced data by varying the $N_{src}/N_{neg}$ ratio by $\pm0.5$.
    
\textbf{Effect of class separation loss:} We carry out the ablation study by removing the class separation loss. We learn the post-increment accuracy of the target domain classifier without applying the class separation loss ($\mathcal{L}_{s1}$). We observe that the average prediction accuracy without the loss minimization was 70.39\% compared to 89.46\% using the class separation loss.

%% file: conclusion.tex
\section{CONCLUSION}

In this work, we present \textbf{ALeN}, an approach for unsupervised domain incremental learning. We present an empirical analysis of our approach by applying it domain incremental learning tasks in sentiment prediction and disease classification applications. We compare our approach to existing state-of-the-art methods to show that the method achieves promising results all our experiments. We present a structured ablation study to observe the impact of different components of our proposed algorithm. Our work shows that we can incrementally update DNN classification models using unlabelled data without storing past training data.

One of the limitations of this method is that it assumes the source and target data to belong to the same global feature space, which might not be the case in several applications. We identify this problem setting as \textit{feature incremental learning}, and aim to address it in our future works. Also, due to limited size of the parametric space of the DNN models (assumed in our work), we observe that our method is prone to overfitting after multiple increments. In this paper, we have not analysed this issue and aim to do so in our future works. Some works in the literature overcome this limitation by adding additional model parameters as new domains are incrementally added \citep{li2017learning, polikar2010learn}.